\documentclass{article}

\usepackage[final]{nips_2016}

\usepackage[utf8]{inputenc} % allow utf-8 input
\usepackage[T1]{fontenc}    % use 8-bit T1 fonts
\usepackage{url}            % simple URL typesetting
\usepackage{graphicx} 
\usepackage{booktabs}       % professional-quality tables
\usepackage{amsfonts}       % blackboard math symbols
\usepackage{nicefrac}       % compact symbols for 1/2, etc.
\usepackage{microtype}      % microtypography
\usepackage{algorithm,algorithmic}

\title{Spikes as Regularizers}

\author{
  Anders S{\o}gaard\thanks{This research is funded by the ERC Starting
Grant LOWLANDS No. 313695, as well as by the Danish Research Council.} \\
  Department of Computer Science\\
  University of Copenhagen\\
  Copenhagen, DK-2200 \\
  \texttt{soegaard@di.ku.dk} 
}

\begin{document}

\maketitle

\begin{abstract}
We present a confidence-based single-layer feed-forward learning algorithm {\sc Spiral}~(Spike Regularized Adaptive Learning) relying on an encoding of activation {\em spikes}. We adaptively update a weight vector relying on confidence estimates and activation offsets relative to previous activity. We regularize updates proportionally to item-level confidence and weight-specific support, loosely inspired by the observation from neurophysiology that high spike rates are sometimes accompanied by low temporal precision. Our experiments suggest that the new learning algorithm {\sc Spiral} is more robust and less prone to overfitting than both the averaged perceptron and {\sc Arow}. 
\end{abstract}

\section{Confidence-weighted Learning of Linear Classifiers}

The perceptron \citep{Rosenblatt:58} is a conceptually simple and widely used discriminative and linear classification algorithm. It was originally motivated by observations of how signals are passed between neurons in the brain. We will return to the perceptron as a model of neural computation, but from a more technical point of view, the main weakness of the perceptron as a linear classifier is that it is prone to overfitting. One particular type of overfitting that is likely to happen in perceptron learning is {\em feature swamping} \citep{Sutton:ea:06}, i.e., that very frequent features may prevent co-variant features from being updated, leading to catastrophic performance if the frequent features are absent or less frequent at test time. In other words, in the perceptron, as well as in passive-aggressive learning \cite{Crammer:ea:06}, parameters are only updated when features occur, and rare features therefore often receive inaccurate values. 

There are several ways to approach such overfitting, e.g., capping the model's supremum norm, but here we focus on a specific line of research: confidence-weighted learning of linear classifiers. Confidence-weighted learning explicitly estimates confidence during induction, often by maintaining Gaussian distributions over parameter vectors. In other words, each model parameter is interpreted as a mean, and augmented with a covariance estimate.  Confidence-Weighted Learning {\sc CWL} \citep{Dredze:ea:08} was the first learning algorithm to do this, but \citet{Crammer:ea:09} later introduced Adaptive Regularization of Weight Vectors ({\sc Arow}), which is a simpler and more effective alternative:

{\sc Arow} passes over the data, item by item, computing a margin, i.e., a dot product of a weight vector $\mathbf{\mu}$ and the item, and updating $\mathbf{\mu}$ and a covariance matrix $\Sigma$ in a standard additive fashion. As in {\sc CWL}, the weights -- which are interpreted as means -- and the covariance matrix form a Gaussian distribution over the weight vectors. Specifically, the confidence is $\mathbf{x}^\top\Sigma\mathbf{x}$. We add a smoothing constant $r(=0.1)$ and compute the learning rate $\alpha$ adaptively:

\begin{equation}\alpha=\frac{\max(0,1-y\mathbf{x}^\top\mathbf{\mu})}{\mathbf{x}^\top\Sigma\mathbf{x}+r}
\end{equation}

We then update $\mathbf{\mu}$ proportionally to $\alpha$, and update the covariance matrix as follows:

\begin{equation} \Sigma\leftarrow \frac{\Sigma-\Sigma\mathbf{x}\mathbf{x}^\top\Sigma}{\mathbf{x}^\top\Sigma\mathbf{x}+r}
\end{equation}

{\sc CWL}~and {\sc Arow}~have been shown to be more robust than the (averaged) perceptron in several studies \citep{Crammer:ea:12,Soegaard:Johannsen:12}, but below we show that replacing binary activations with samples from spikes can lead to better regularized and more robust models. 

\section{Spikes as Regularizers}

\subsection{Neurophysiological motivation}

Neurons do not fire synchronously at a constant rate. Neural signals are spike-shaped with an onset, an increase in signal, followed by a spike and a decrease in signal, and with an inhibition of the neuron before returning to its equilibrium. Below we simplify the picture a bit by assuming that spikes are bell-shaped (Gaussians). 

The learning algorithm ({\sc Spiral}) which we will propose below, is motivated by the observation that spike rate (the speed at which a neuron fires) increases the more a neuron fires \citep{Kawai:Sterling:02,Keller:Takahashi:15}. Futhermore, \citet{Keller:Takahashi:15} show that increased activity may lead to spiking at higher rates with lower temporal precision. This means that the more active neurons are less successful in passing on signals, leading the neuron to return to a more stable firing rate. In other words, the brain performs implicit regularization by exhibiting low temporal precision at high spike rates. This prevents highly active neurons from {\em swamping}~other co-variant, but less active neurons. We hypothesise that implementing a similar mechanism in our learning algorithms will prevent feature swamping in a similar fashion. 

Finally, \citet{Blanco:ea:15} show that periods of increased spike rate lead to a smaller standard deviation in the synaptic weights. This loosely inspired us to implement the temporal imprecision at high spike rates by decreasing the weight's standard deviation. 

\subsection{The algorithm}

In a single layer feedforward model, such as the perceptron, sampling from Gaussian spikes only effect the input, and we can therefore implement our regularizer as noise injection \citep{Bishop:95}. The variance is the relative confidence of the model on the input item (same for all parameters), and the means are the parameter values. We multiply the input by the inverse of the sample, reflecting the intuition that highly active neurons are less precise and more likely to drop out, before we clip the sample from 0 to 1. 

We give the pseudocode in Algorithm~\ref{spiral}, following the conventions in \citet{Crammer:ea:09}. 

\begin{algorithm}
{\footnotesize
 % \label{alg:beast}
  \begin{algorithmic}[1]
%    \Function{Training}{}
    \STATE {$r=0.1, \mathbf{\mu}_0=\mathbf{0},\Sigma_0=I, \{\mathbf{x}_t\mid\mathbf{x}_t\in\mathbb{R}^d\},v=0$}
    \FOR {$t< T$}
    	\STATE {$\mathbf{x}_t=\mathbf{x}_t\cdot $}
	\IF {$v_t>v$}
		\STATE {$v=v_t$}	
	\ENDIF
	\STATE {$v_t=\mathbf{x}_t^\top\Sigma_{t-1}\mathbf{x}_t$ (sampling activations from Gaussian spikes)}
	\STATE {$v_t= 0 < v_t < 1$ (clipping values outside the [0,1] window)}
	\STATE {$\mathbf{\nu}_t\sim \mathcal{N}(\mathbf{\mu}_{t-1},\frac{v_t}{v})$}
	\STATE {$\mathbf{x}_t=\mathbf{x}_t\cdot(1-\mathbf{\nu}_t)$}
	\STATE {$m_t=\mathbf{\mu}_{t-1}\cdot\mathbf{x}_t$}
	\IF {$m_ty_t<1$}
		\STATE {$\alpha_t=\frac{\max(0,1-y_t\mathbf{x}_t^\top\mathbf{\mu}_{t-1})}{\mathbf{x}_t^\top\Sigma_{t-1}\mathbf{x}_t+r}$}
	\STATE {$\mathbf{\mu}_t=\mathbf{\mu}_{t-1}+\alpha_t\Sigma_{t-1}y_t\mathbf{x}_t$}
	\STATE {$\Sigma_t = \frac{\Sigma_{t-1}-\Sigma_{t-1}\mathbf{x}_t\mathbf{x}_t^\top\Sigma_{t-1}}{\mathbf{x}_t^\top\Sigma\mathbf{x}_t+r}$}
	\ENDIF
    \ENDFOR
    \RETURN {$\mathbf{\mu}_T,\Sigma_T$}
  \end{algorithmic}
  \caption{\label{spiral}{\sc Spiral}. Except for lines 8--10, this is identical to {\sc Arow}.}
  }
\end{algorithm}

\begin{figure}
\begin{center}
   \includegraphics[natwidth=4in,width=0.4\linewidth]{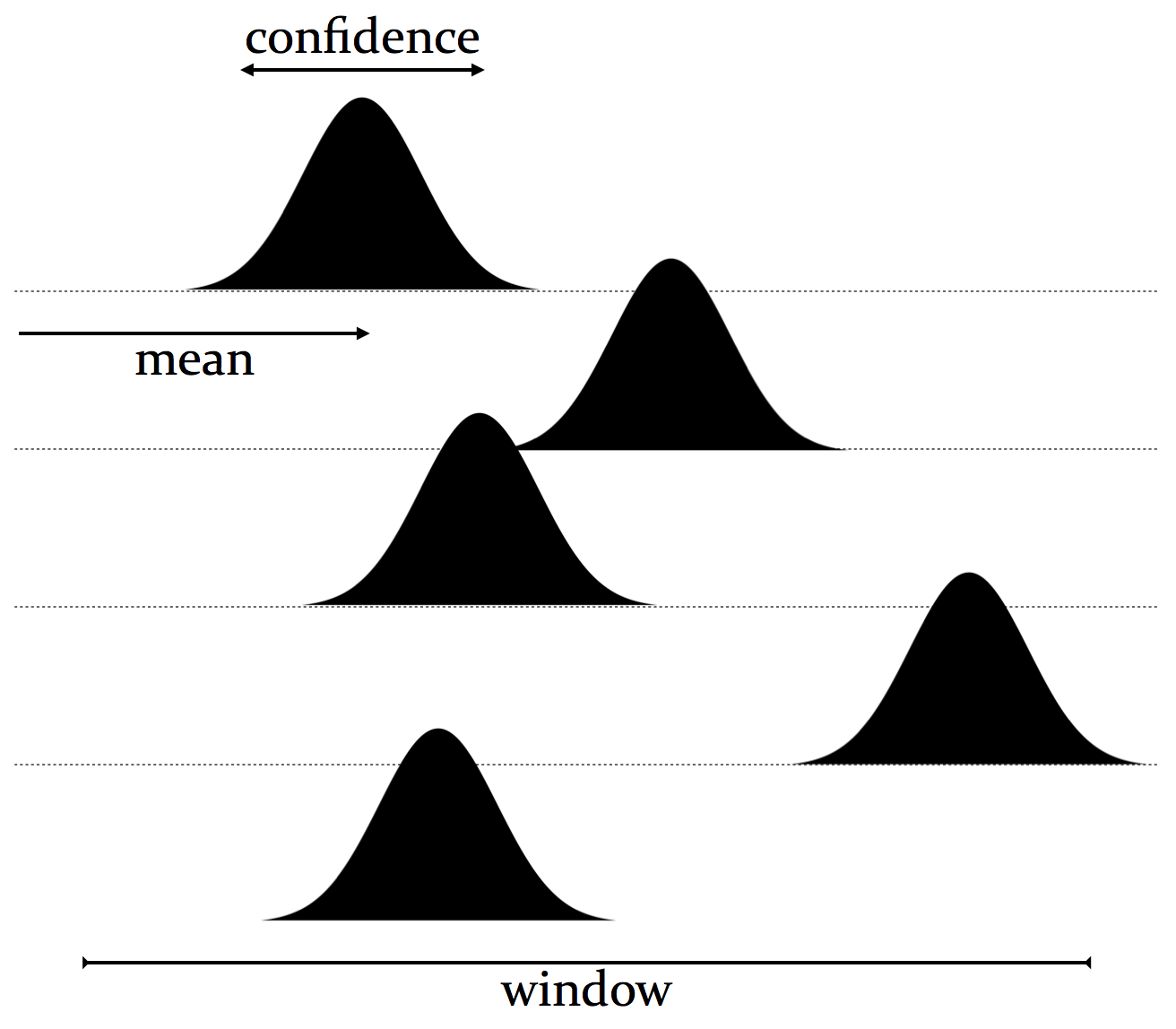}
\caption{Sampling activations from Gaussian spikes.}
\end{center}
\end{figure}

\section{Experiments}

\subsection{Main experiments}

\begin{figure}
\begin{center}
 {\sc Arow}\\
   \includegraphics[width=0.18\linewidth]{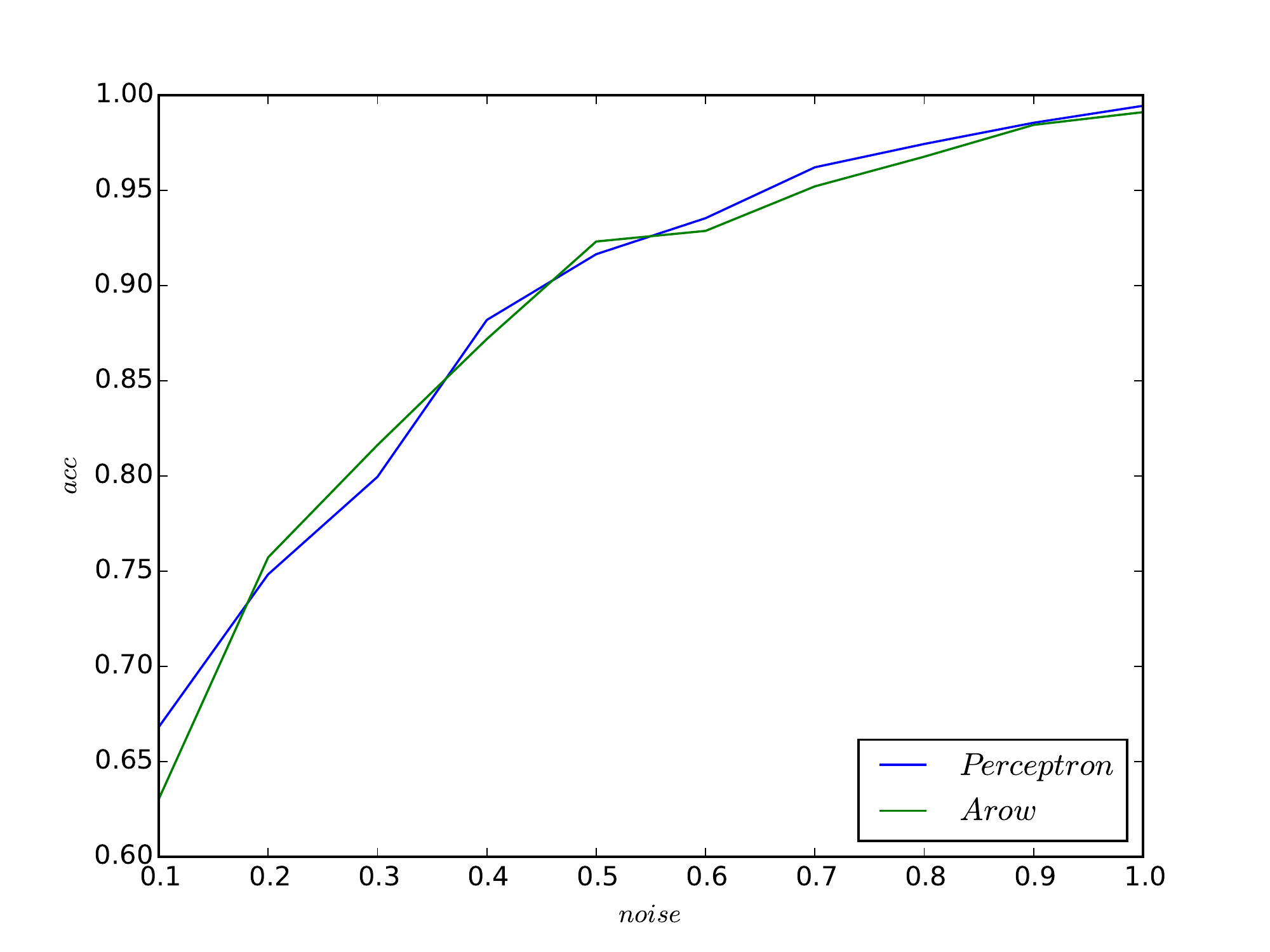}
   \includegraphics[width=0.18\linewidth]{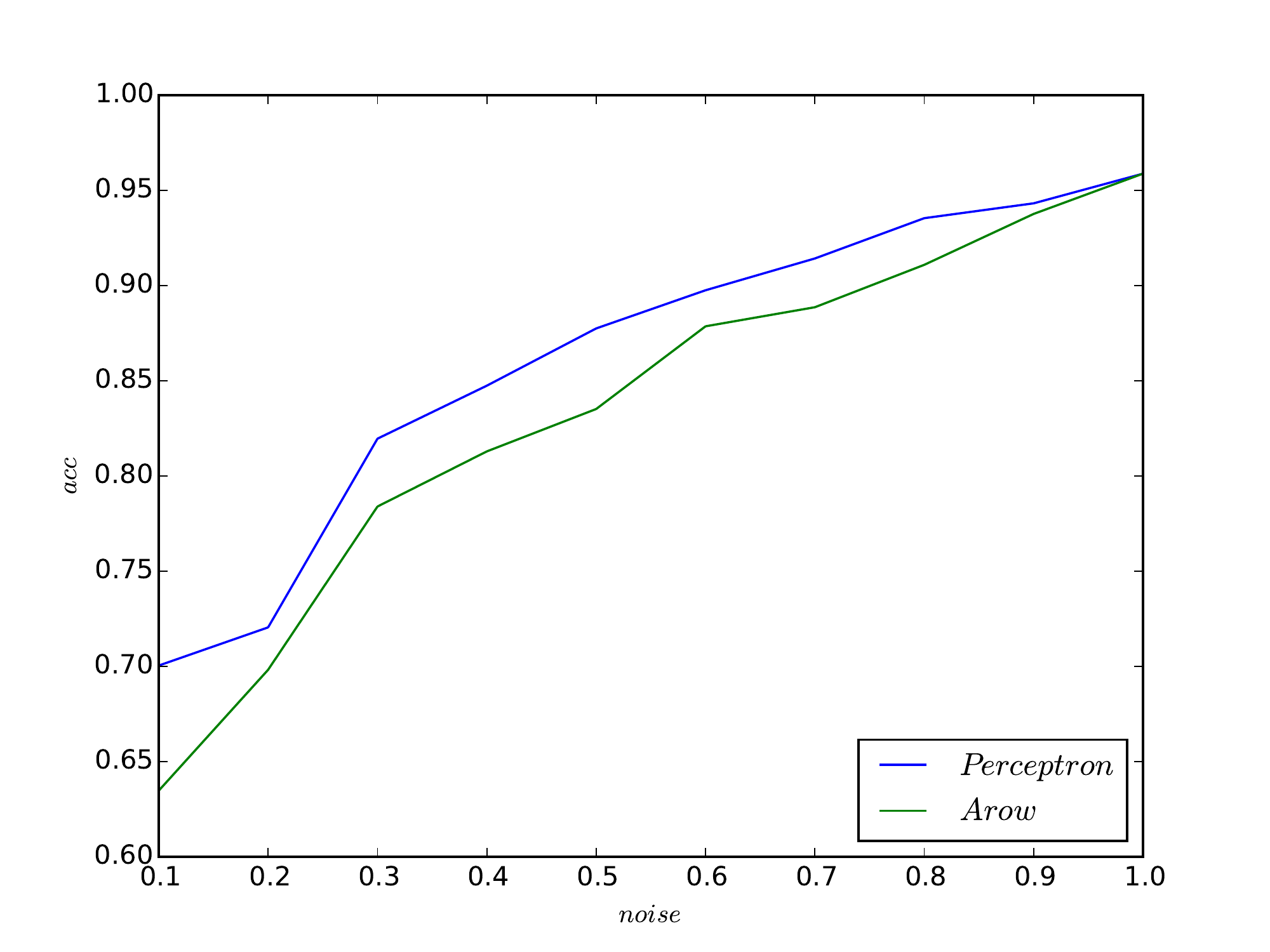}
   \includegraphics[width=0.18\linewidth]{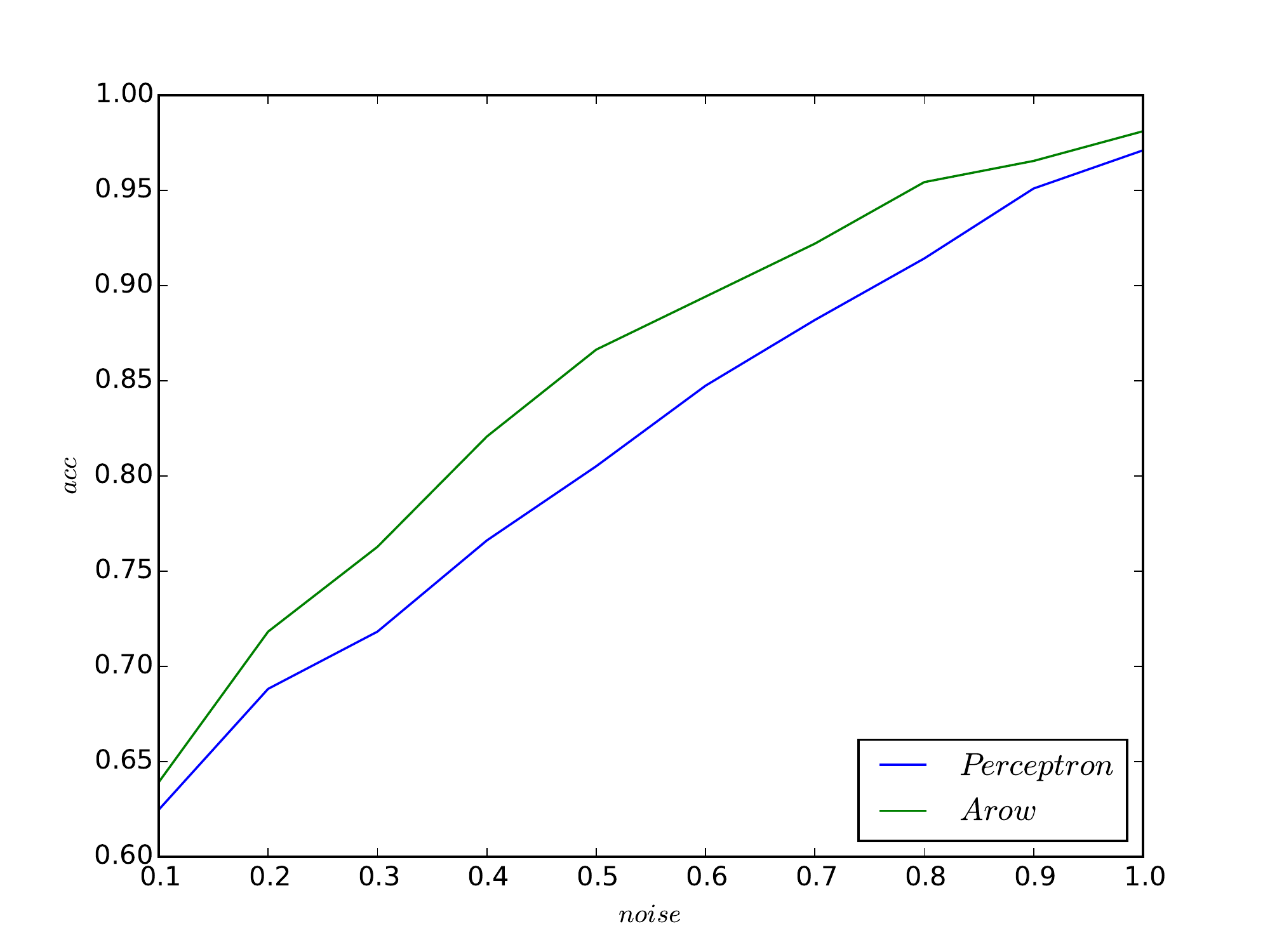}
   \includegraphics[width=0.18\linewidth]{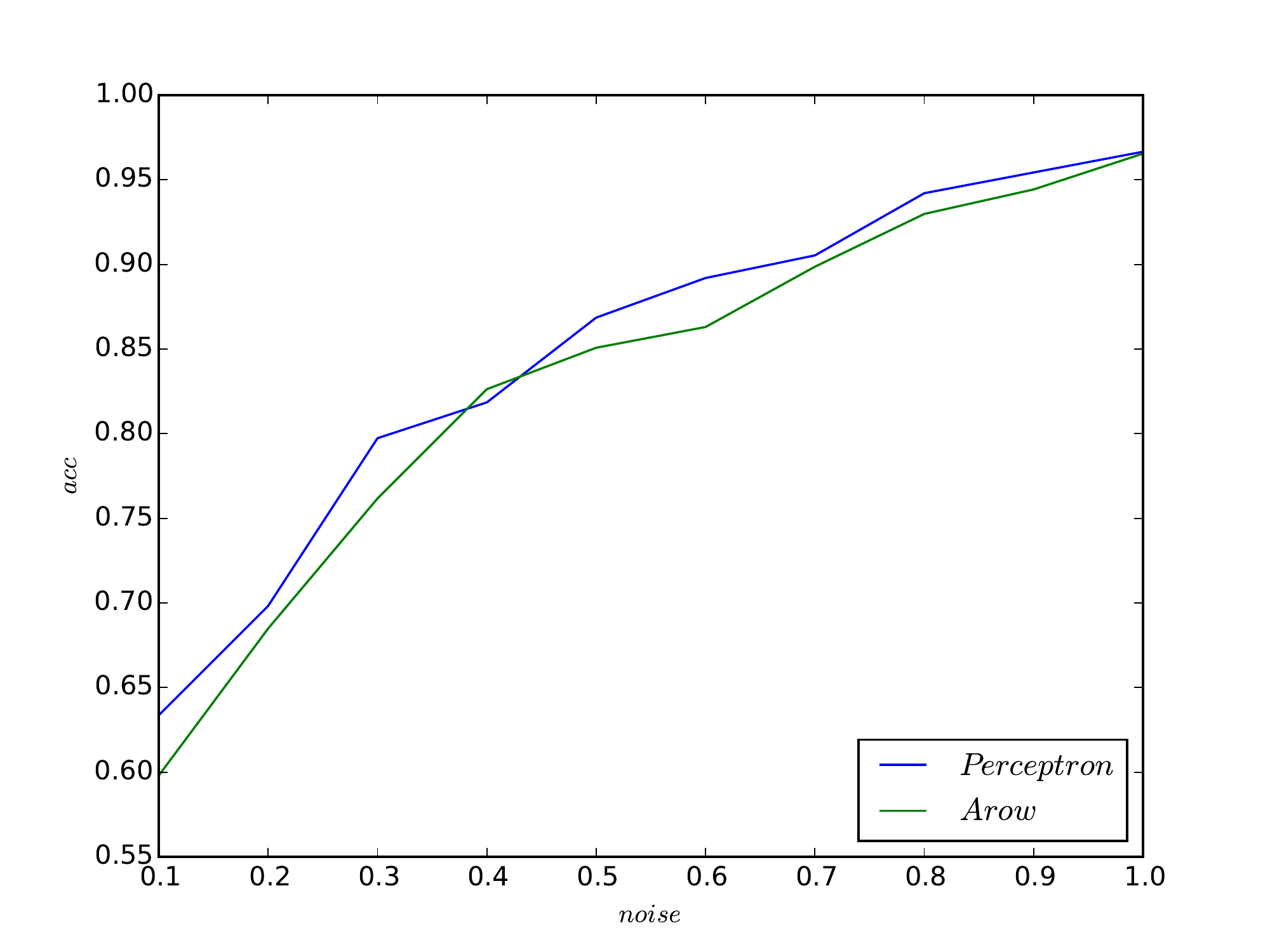}
   \includegraphics[width=0.18\linewidth]{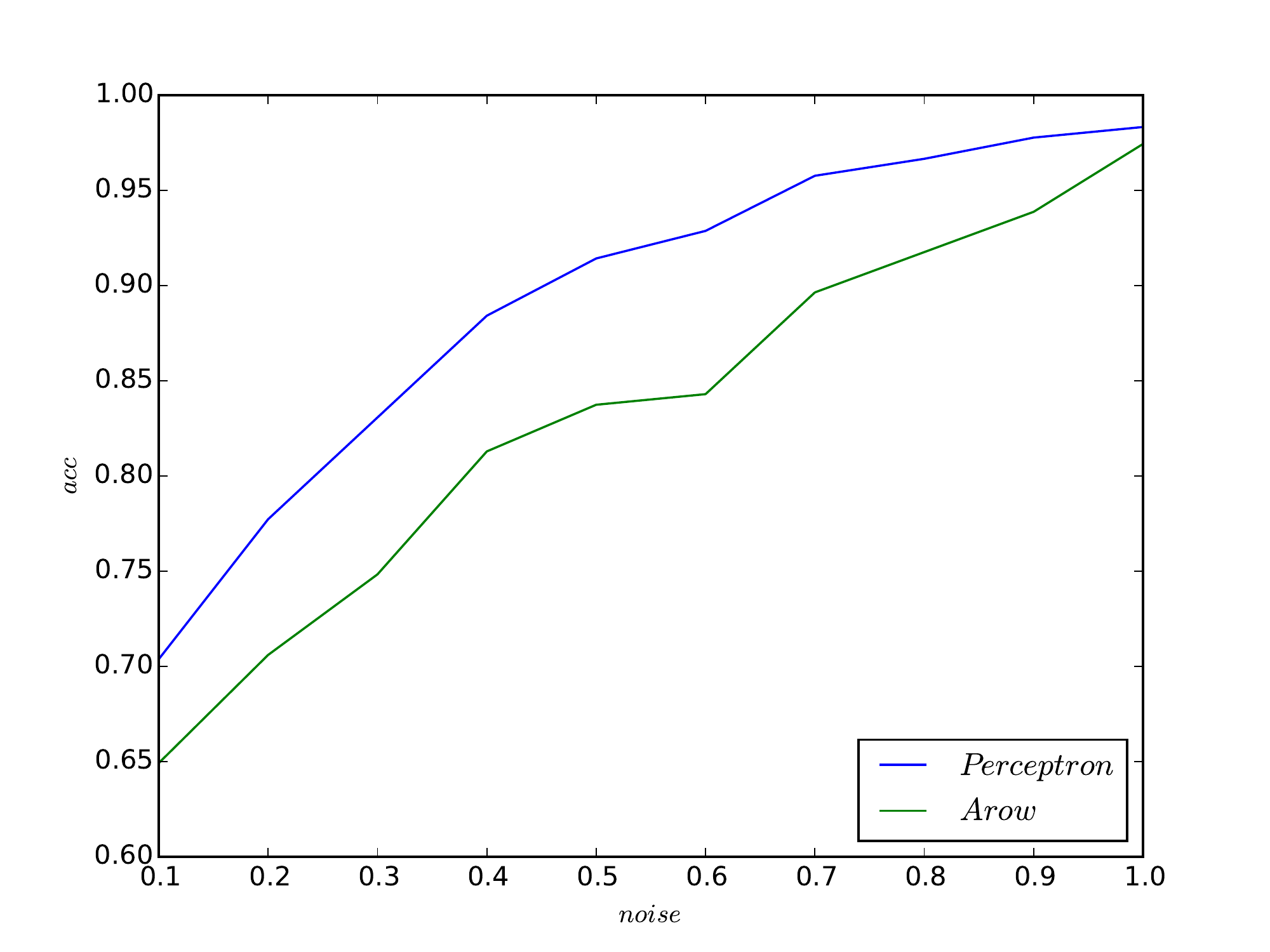}
   \includegraphics[width=0.18\linewidth]{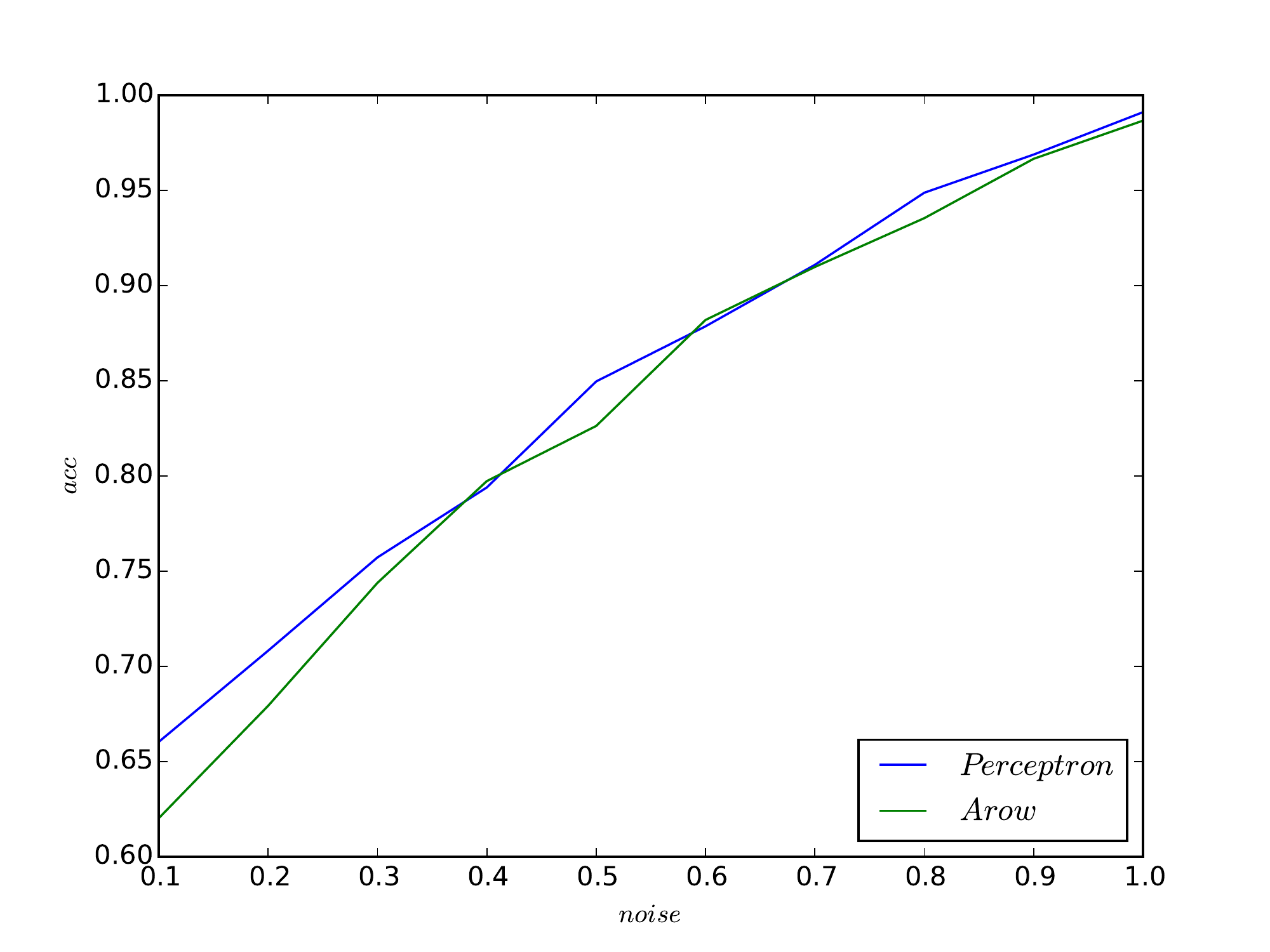}
   \includegraphics[width=0.18\linewidth]{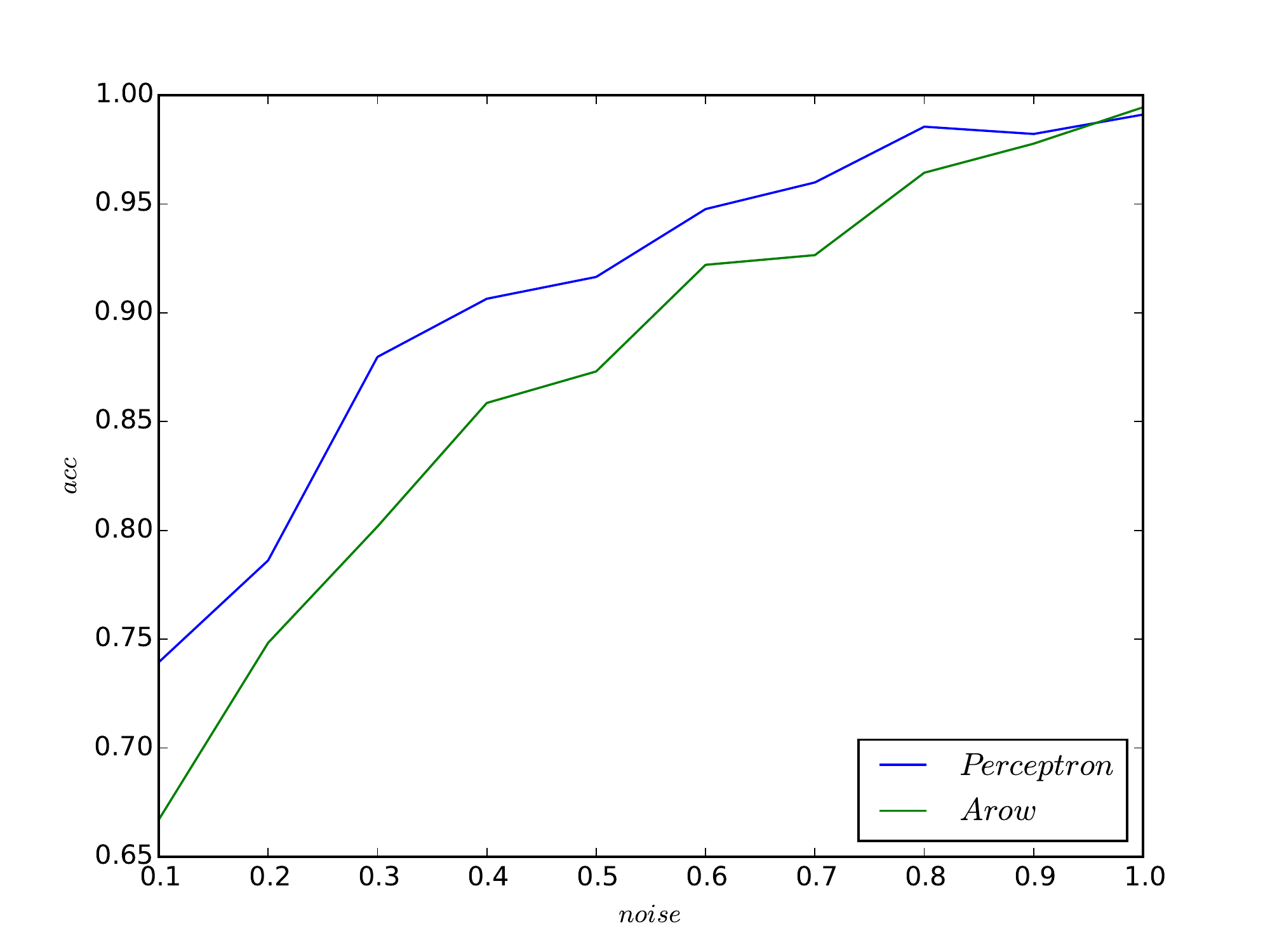}
   \includegraphics[width=0.18\linewidth]{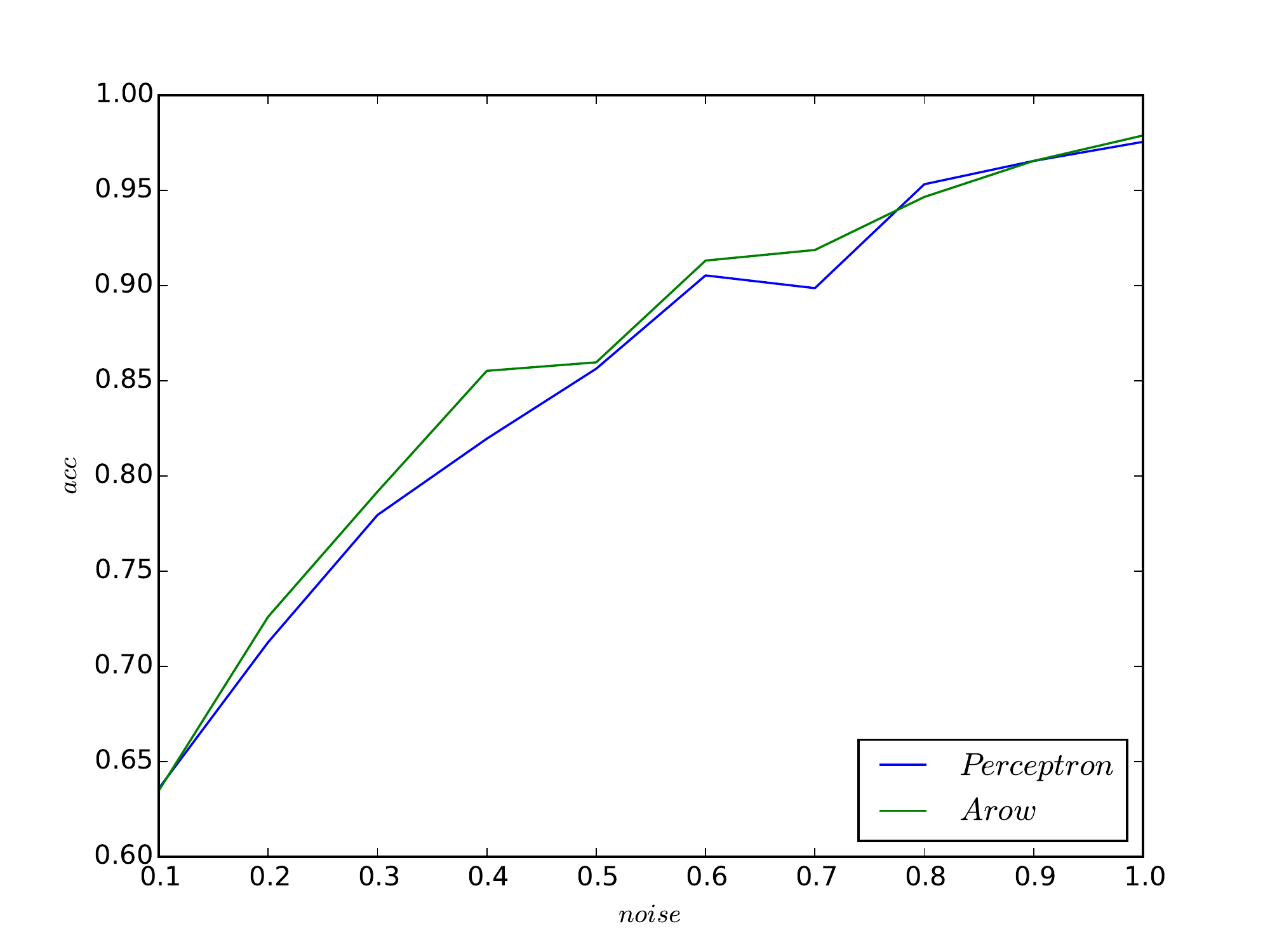}
   \includegraphics[width=0.18\linewidth]{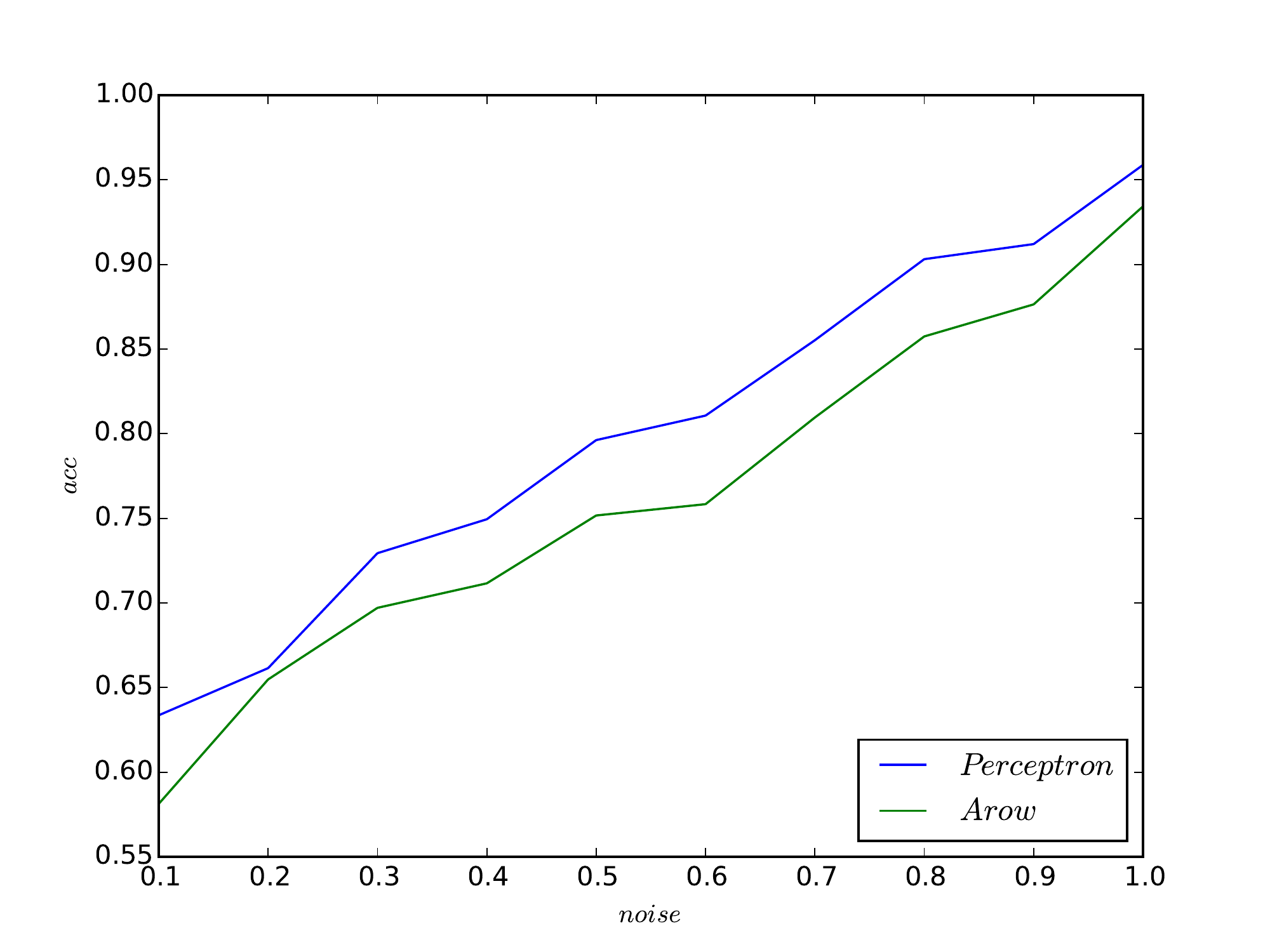}
   \includegraphics[width=0.18\linewidth]{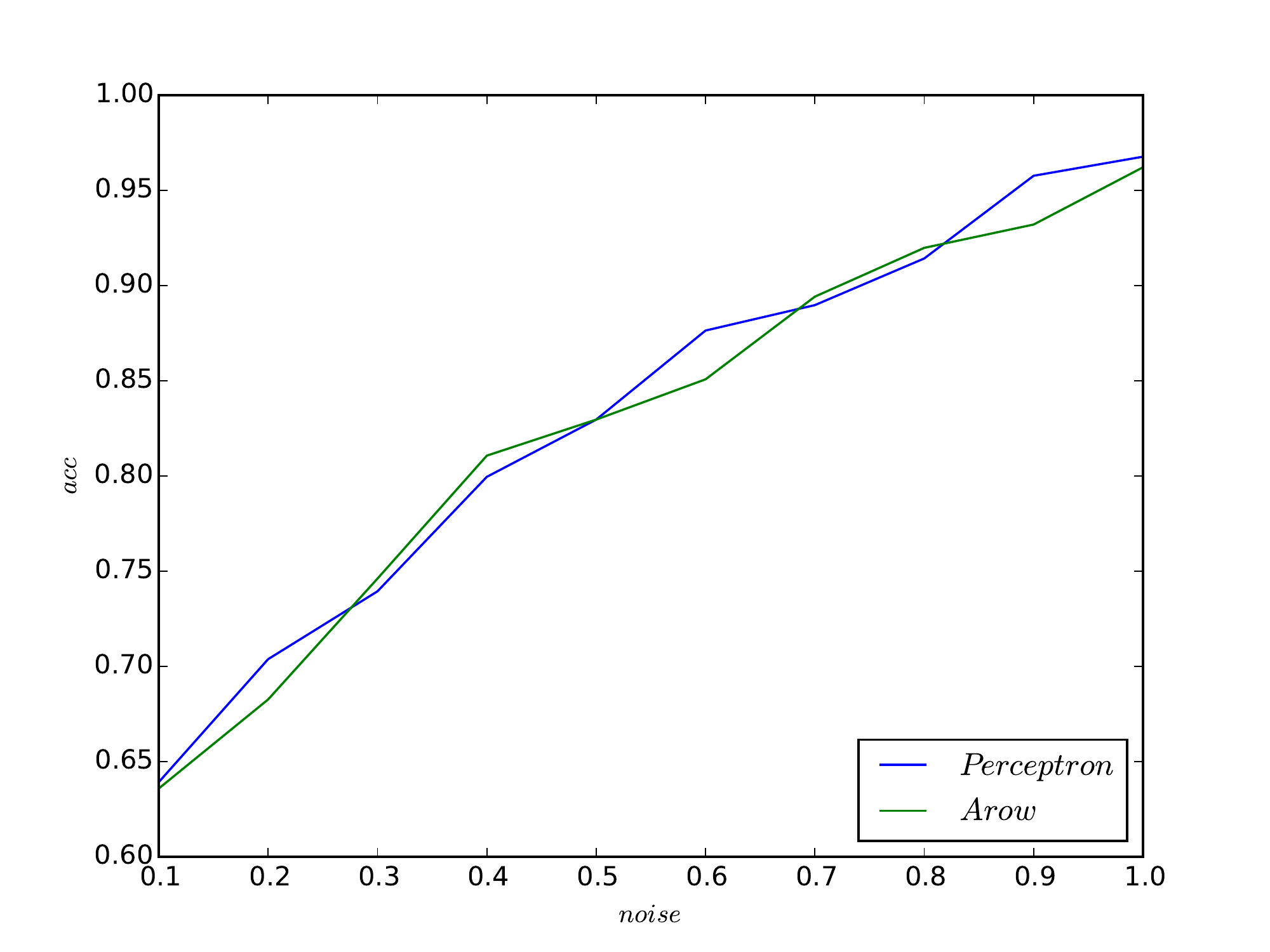}
\\   {\sc Spiral}\\
   \includegraphics[width=0.18\linewidth]{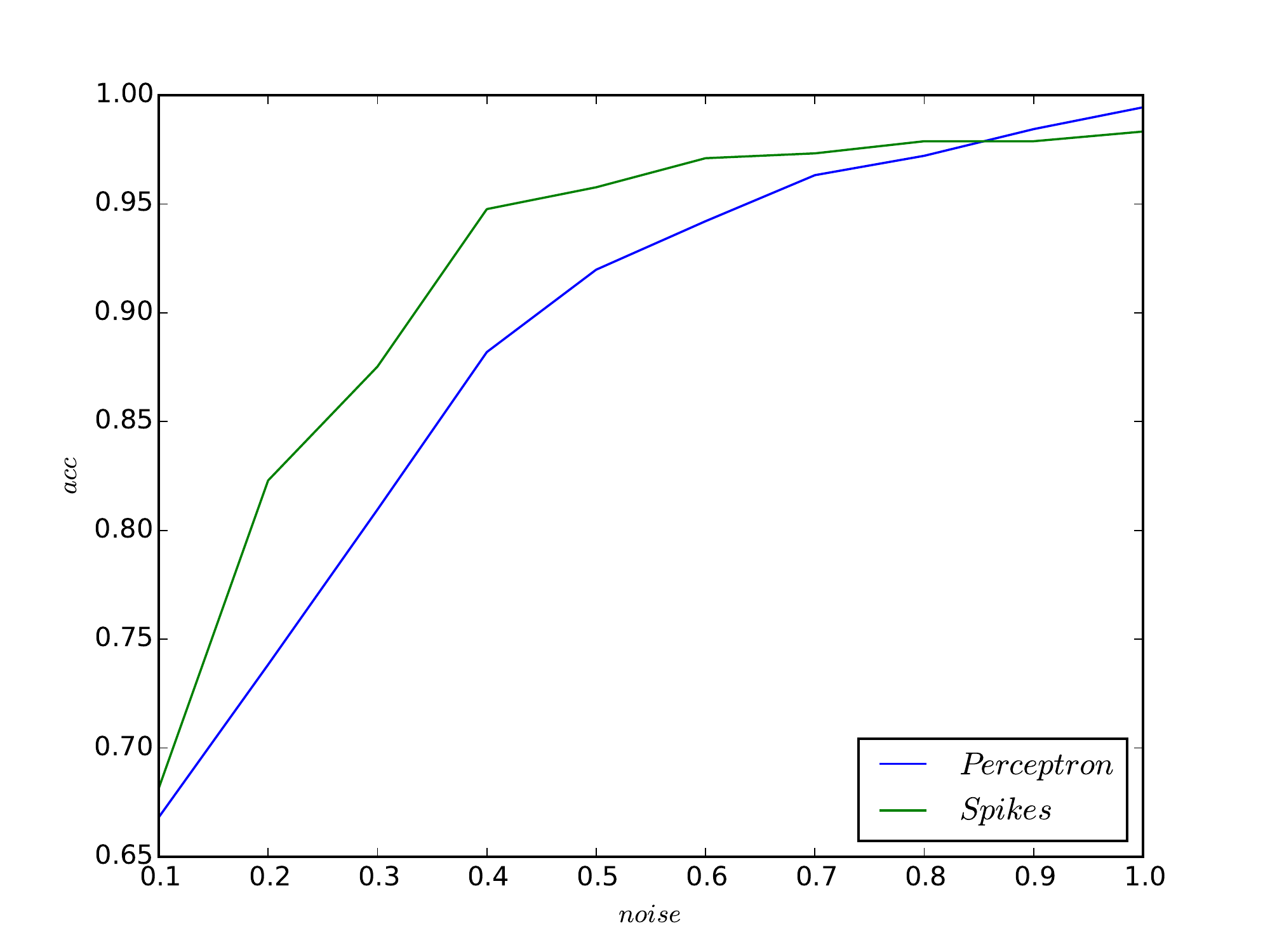}
   \includegraphics[width=0.18\linewidth]{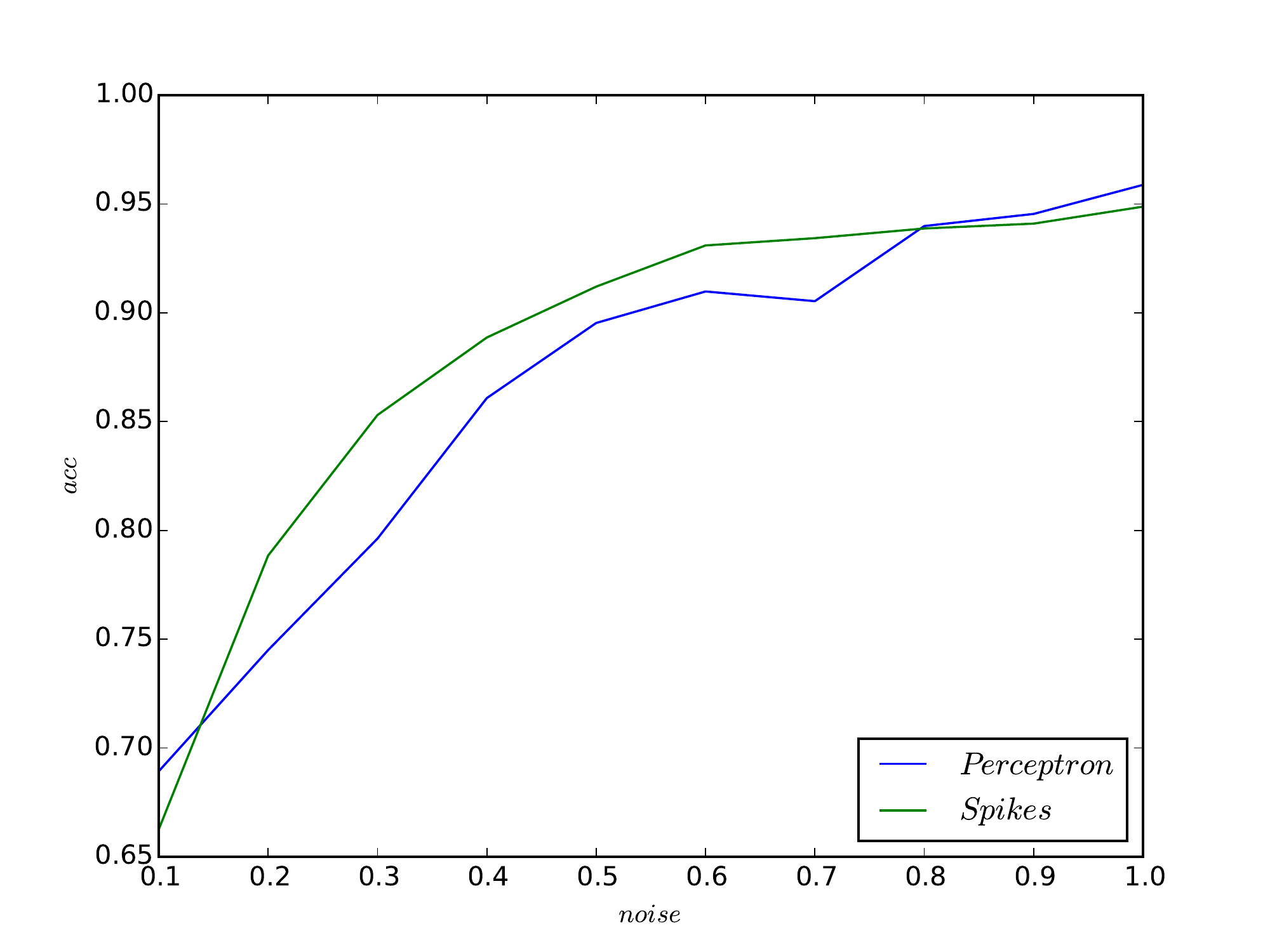}
   \includegraphics[width=0.18\linewidth]{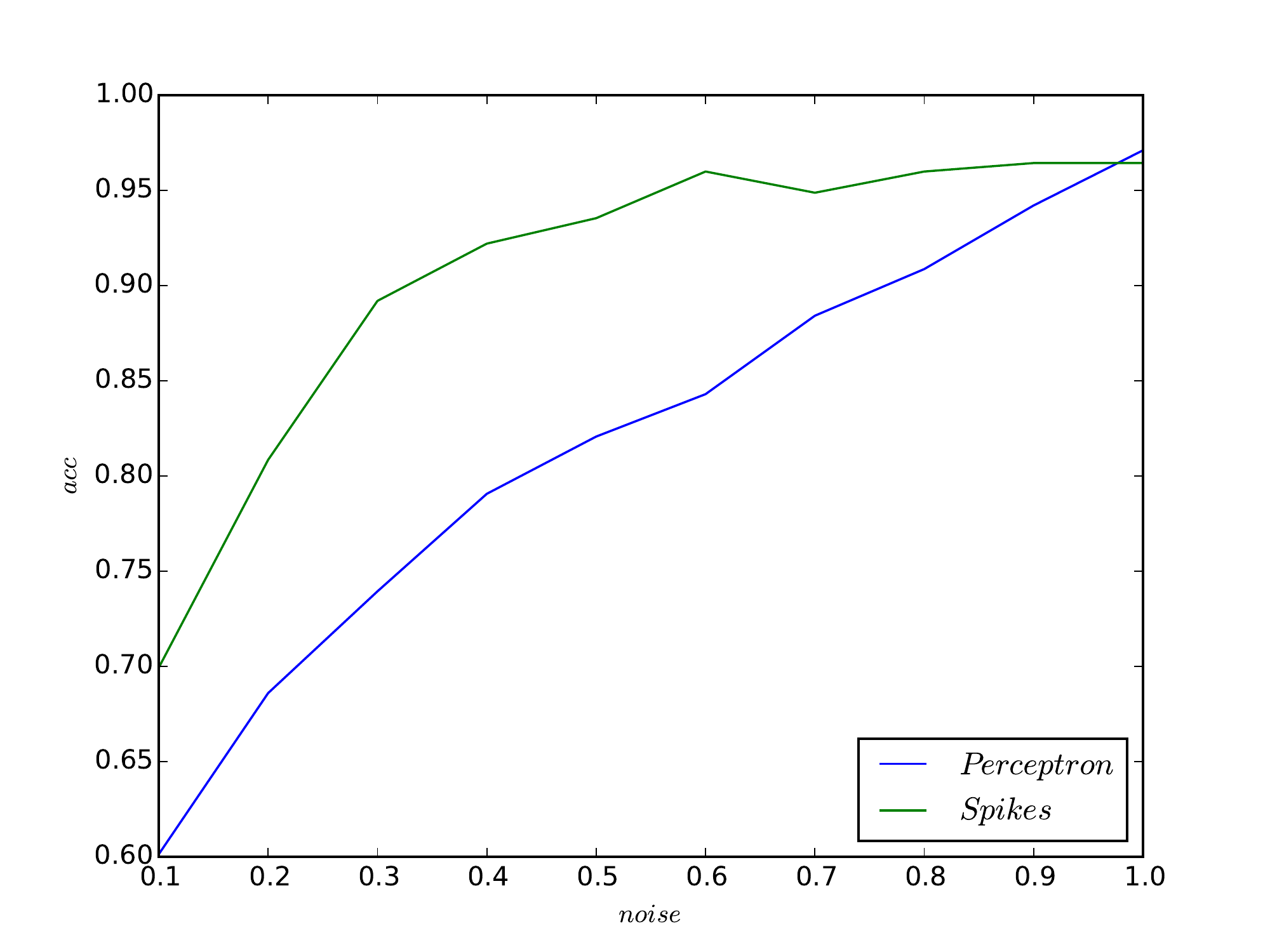}
   \includegraphics[width=0.18\linewidth]{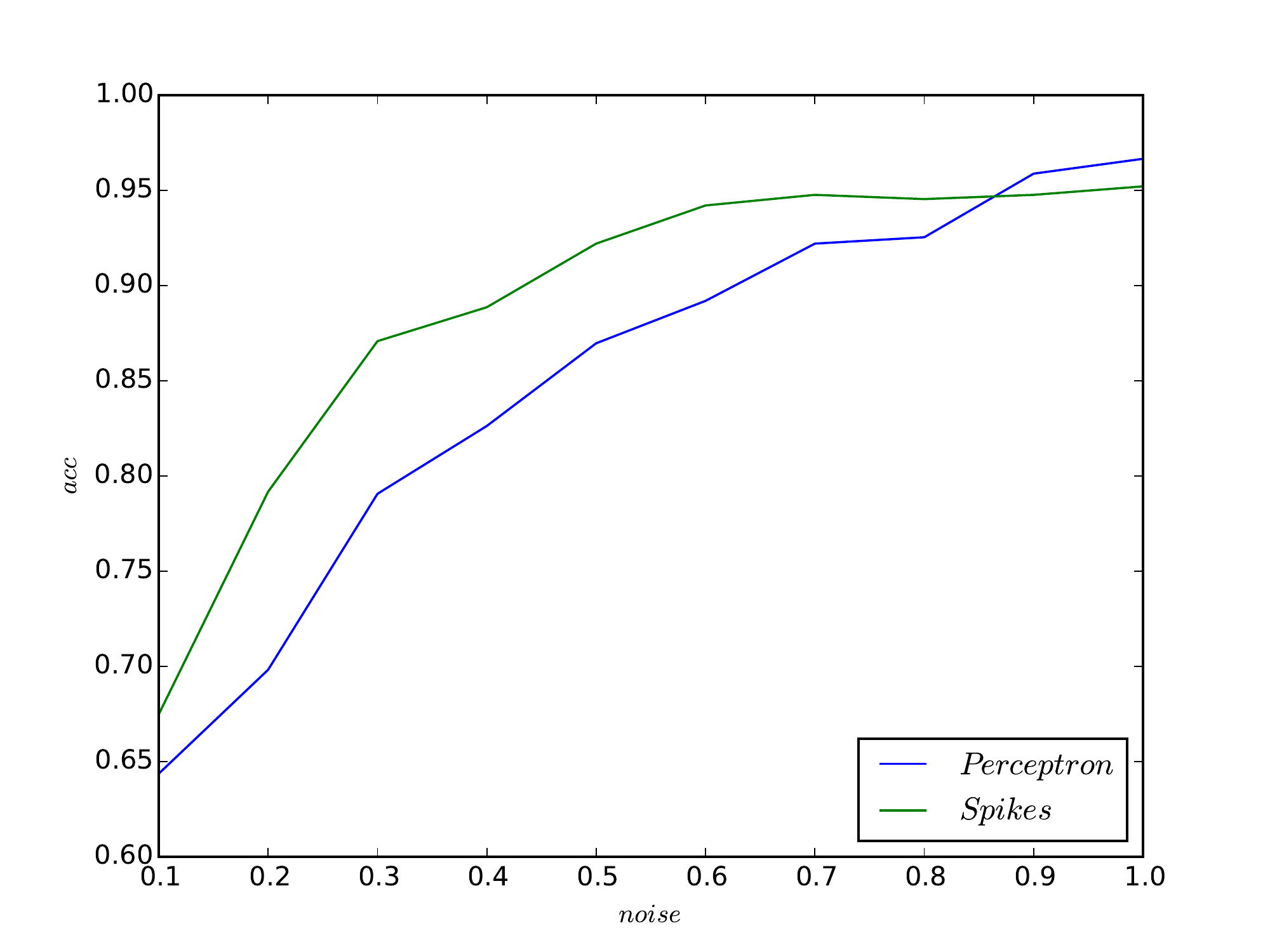}
   \includegraphics[width=0.18\linewidth]{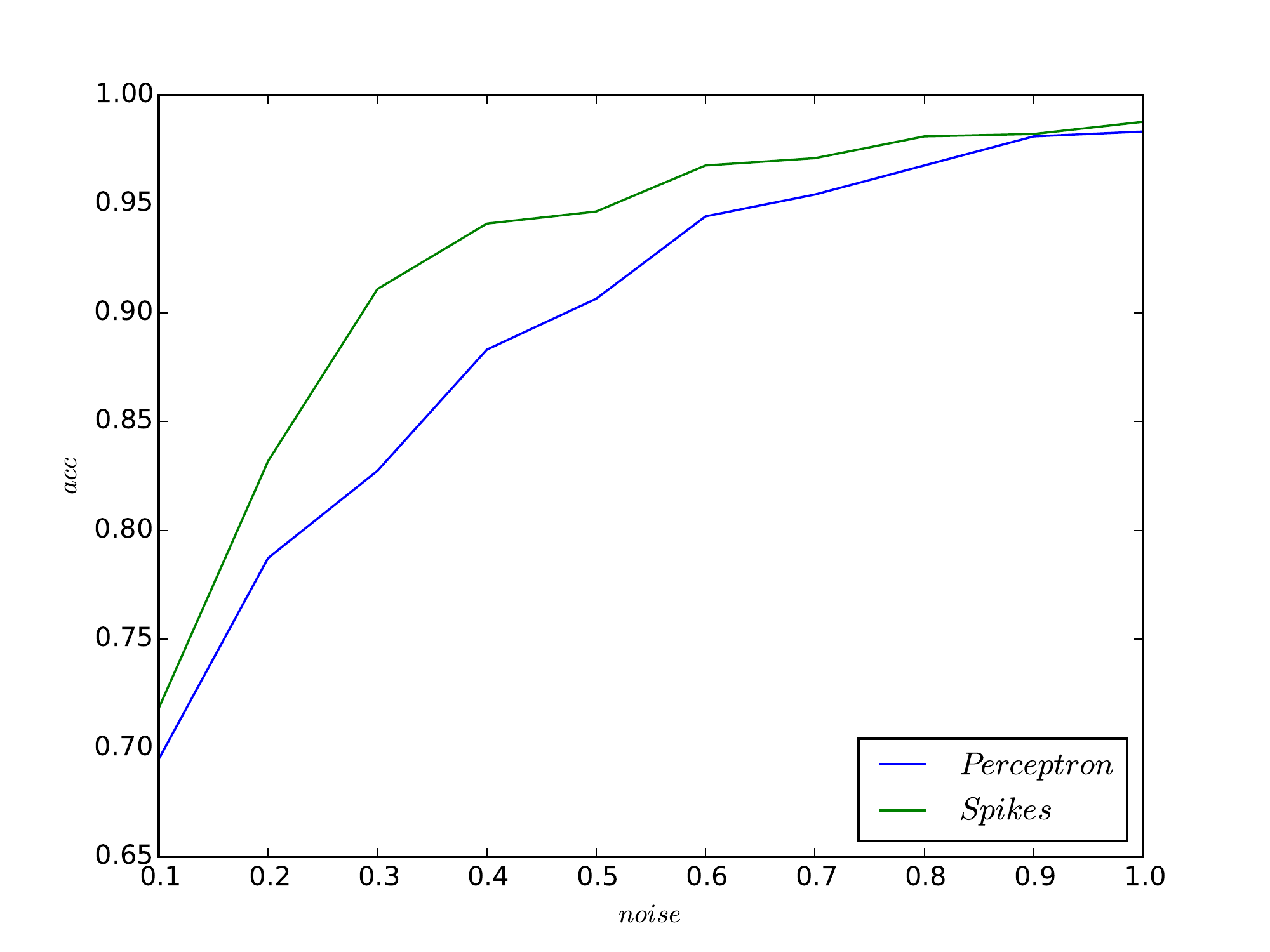}
   \includegraphics[width=0.18\linewidth]{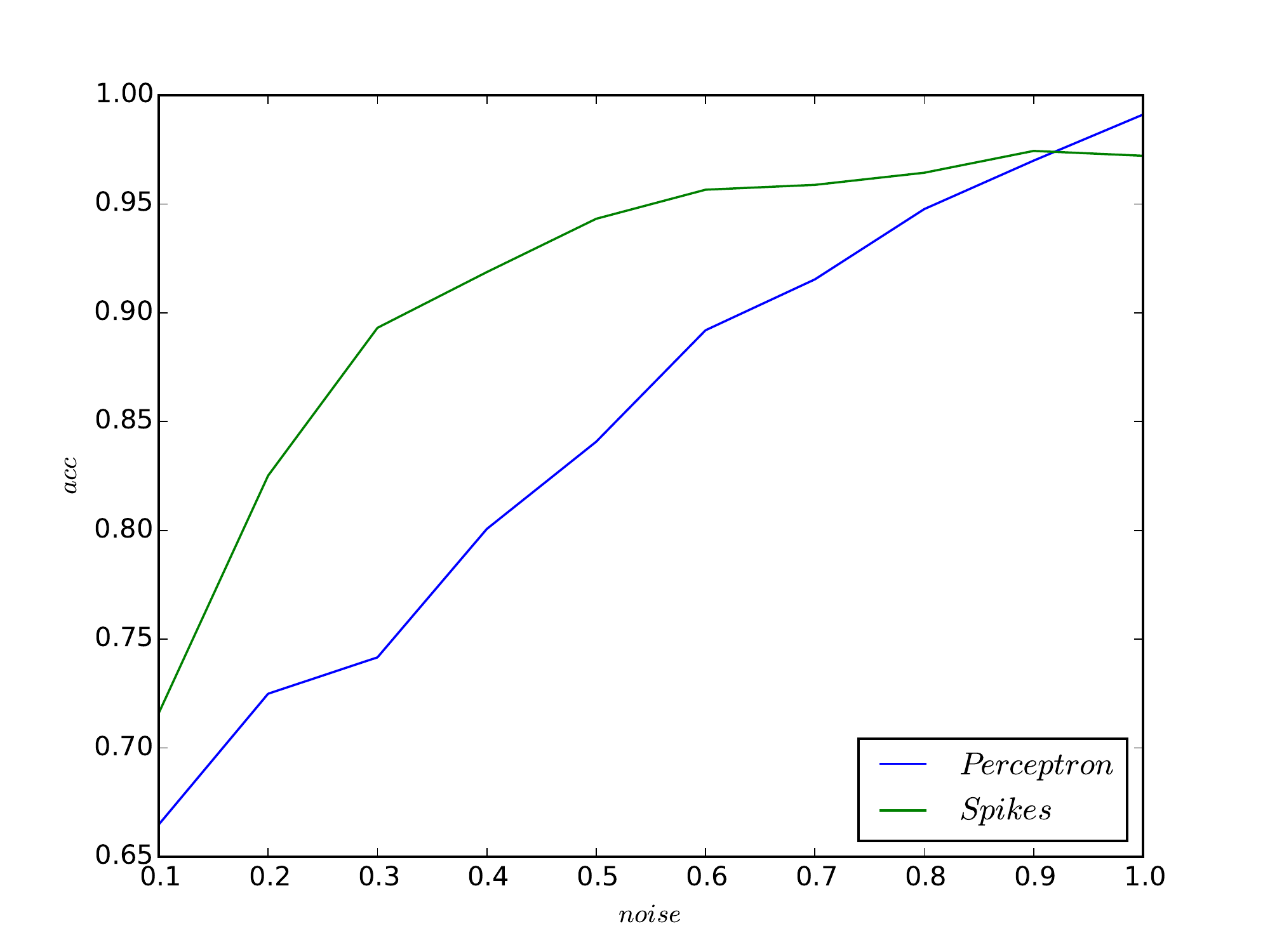}
   \includegraphics[width=0.18\linewidth]{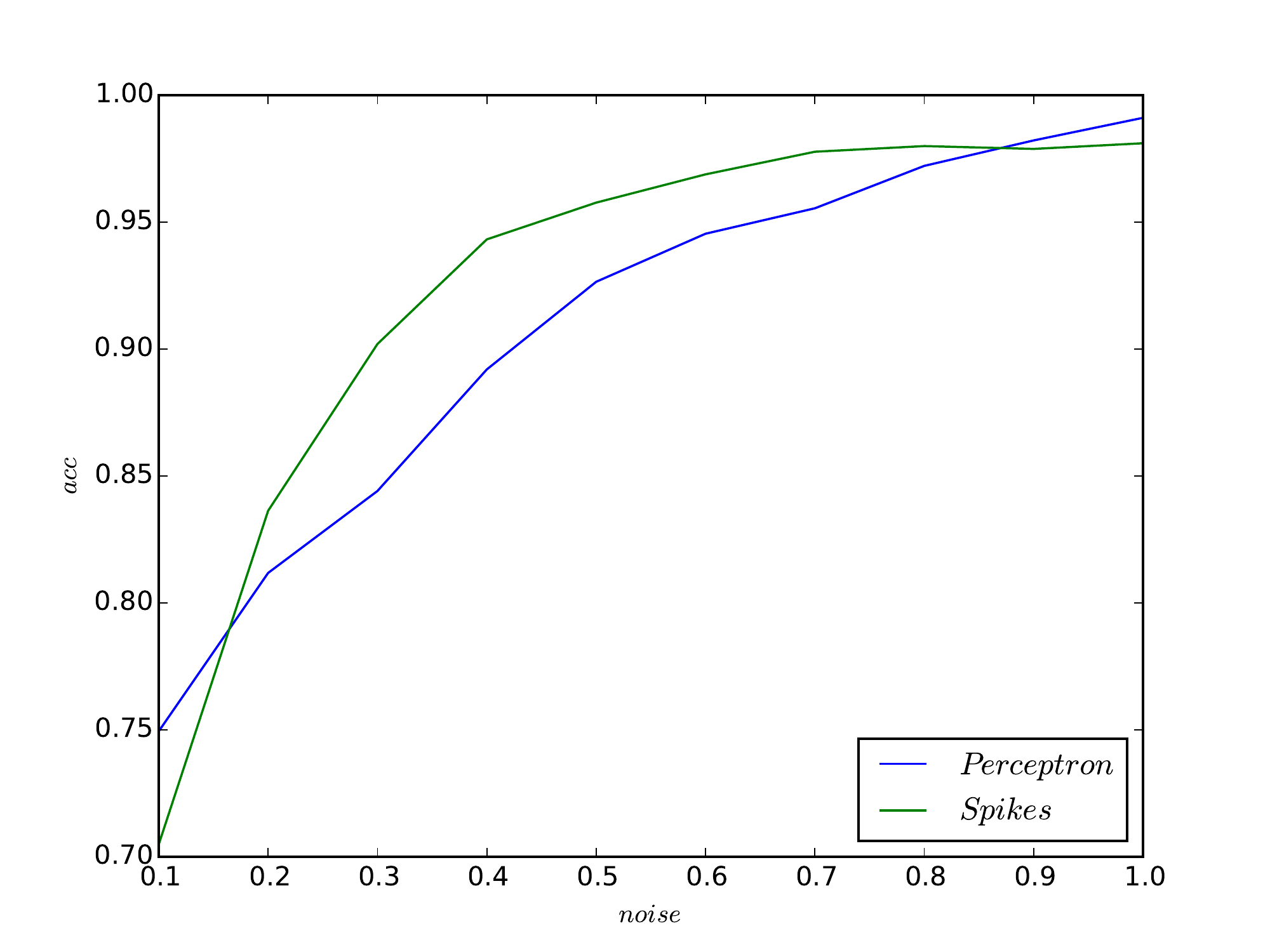}
   \includegraphics[width=0.18\linewidth]{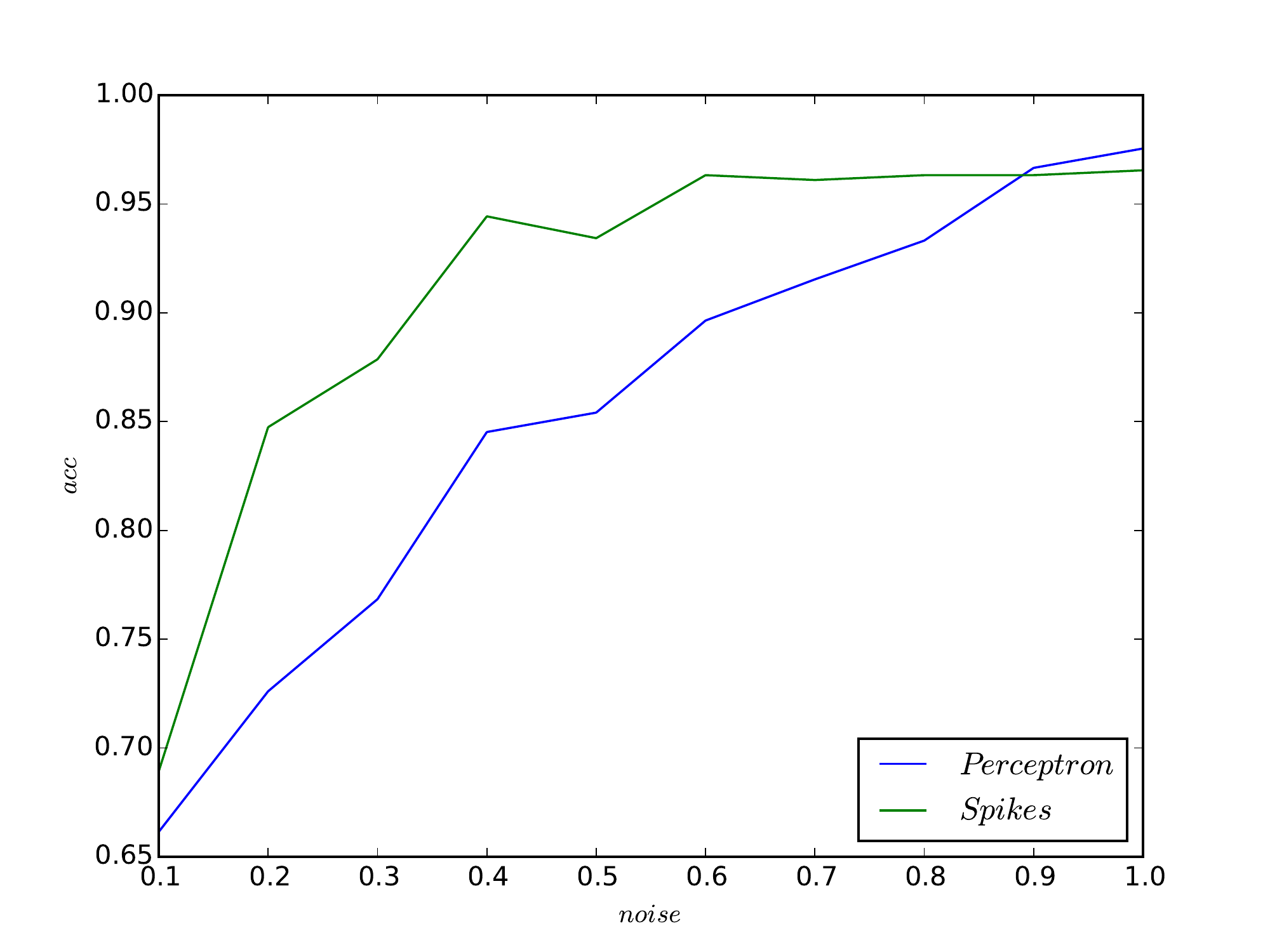}
   \includegraphics[width=0.18\linewidth]{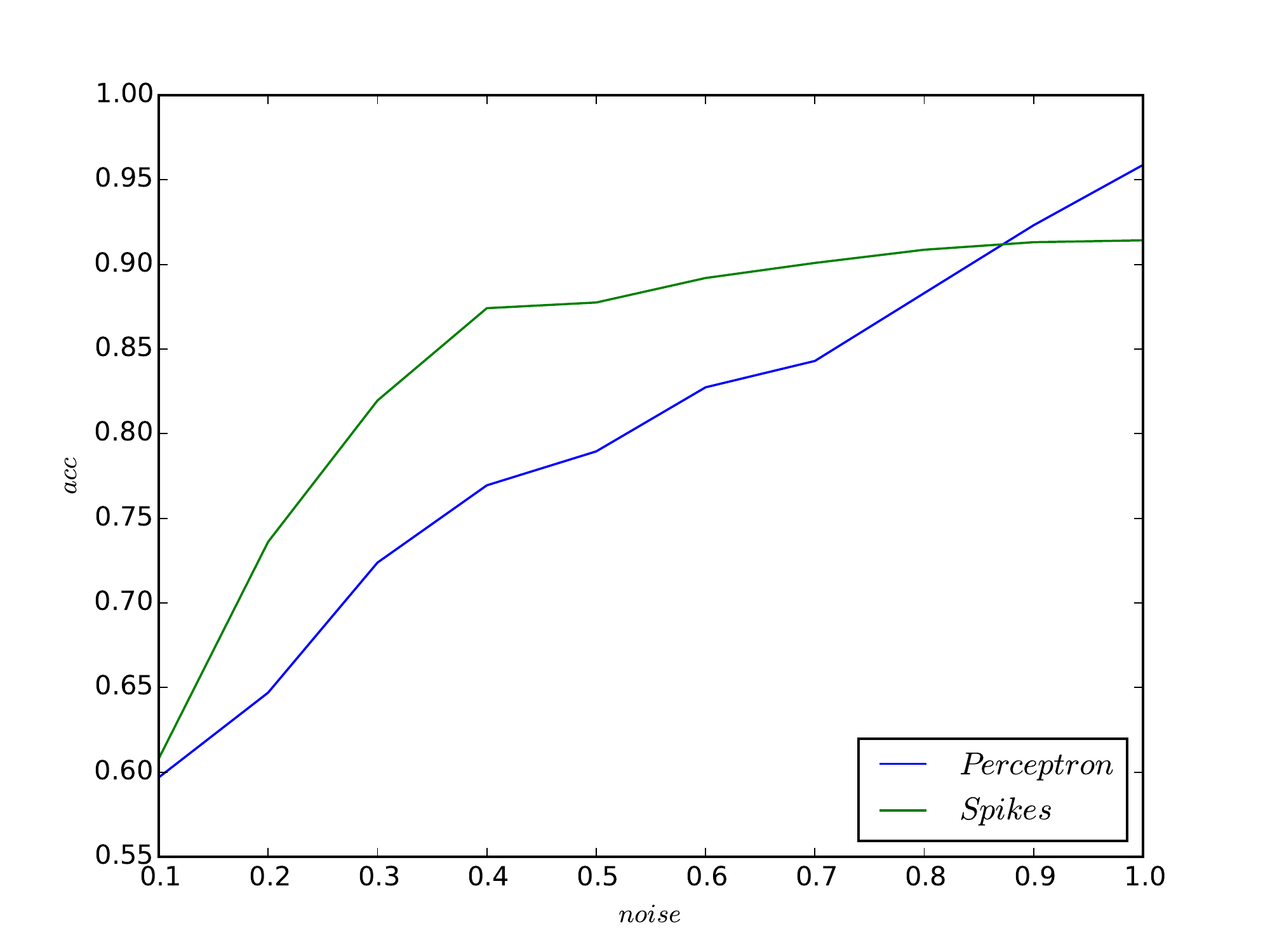}
   \includegraphics[width=0.18\linewidth]{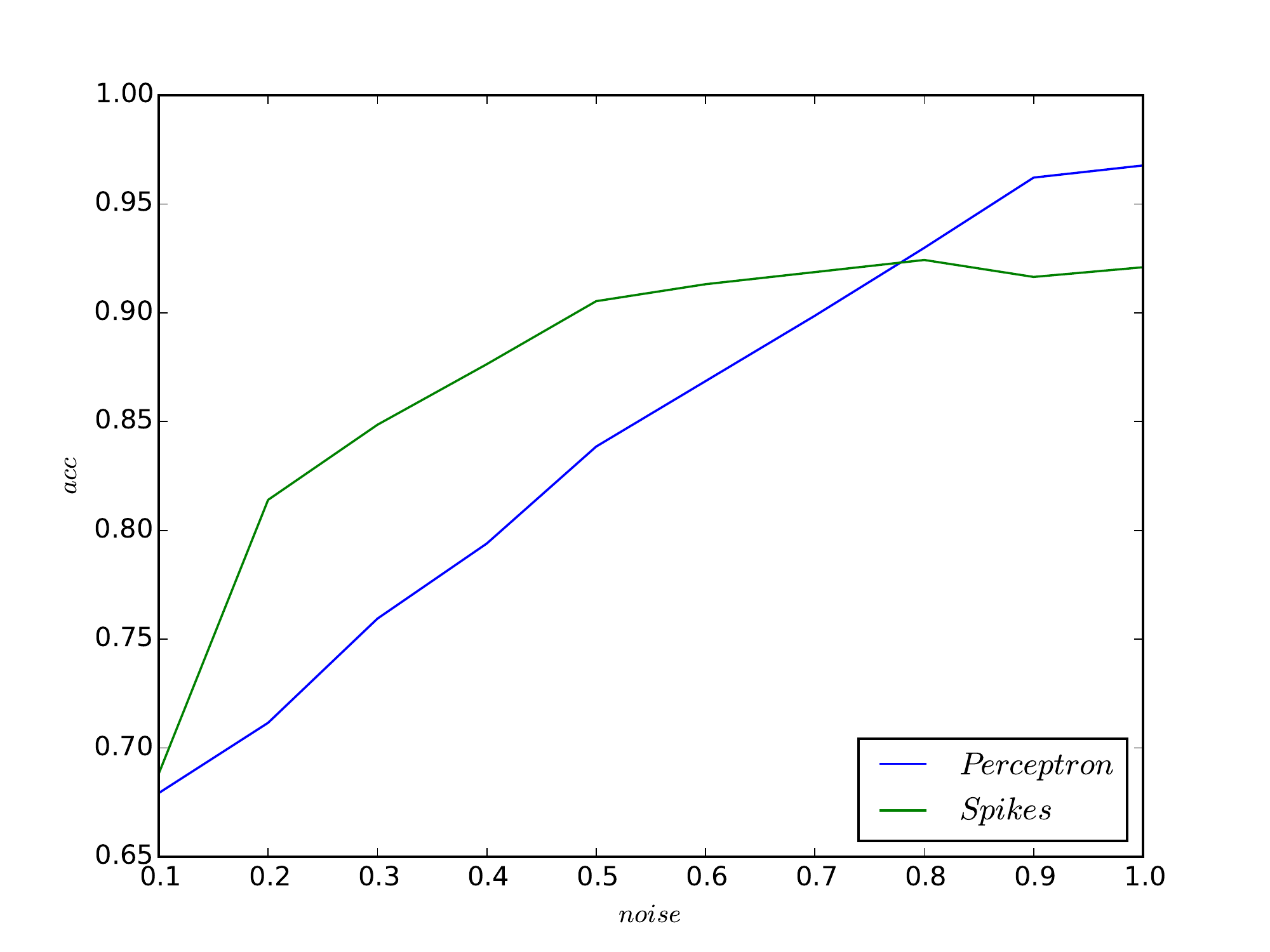}

\caption{\label{plots}Performance over noise levels (percentage of features {\em kept}). First two lines compare the perceptron (blue) and {\sc Arow} (green); the third and fourth compare the perceptron (blue) and {\sc Spiral} (green). }
\end{center}
\end{figure}

We extract 10 binary classification problems from MNIST, training on odd data points, testing on even ones. Since our algorithm is parameter-free, we did not do explicit parameter tuning, but during the implementation of {\sc Spiral}, we only experiment with the first of these ten problems (left, upper corner). To test the robustness of {\sc Spiral} relatively to the perceptron and {\sc Arow}, we randomly corrupt the input at test time by removing features. Our set-up is inspired by \citet{Globerson:Roweis:06}. In the plots in Figure~\ref{plots}, the $x$-axis presents the number of features kept ({\em not} deleted). 

We observe two tendencies in the results: (i) {\sc Spiral}~outperforms the perceptron consistently with up to 80\%~of the features, and sometimes by a very large margin; except that in 2/10 cases, the perceptron is better with only 10\%~of the features. (ii) In contrast, {\sc Arow}~is less stable, and only improves significantly over the perceptron under mid-range noise levels in a few cases. The perceptron is almost always superior on the full set of features, since this is a relatively simple learning problem, where overfitting is unlikely, unless noise is injected at test time. 

\subsection{Practical Rademacher complexity}

We compute {\sc Spiral}'s practical Rademacher complexity as the ability of {\sc Spiral}~to fit random re-labelings of data. We randomly label the above dataset ten times and compute the average error reduction over a random baseline. The perceptron achieves a 5\%~error reduction over a random baseline, on average, overfitting quite a bit to the random labelling of the data. In contrast, {\sc Spiral}~only reduces 0.6\%~of the errors of a random baseline on average, suggesting that it is almost resilient to overfitting on this dataset. 

\section{Conclusion}

We have presented a simple, confidence-based single layer feed-forward learning algorithm {\sc Spiral}~that uses sampling from Gaussian spikes as a regularizer, loosely inspired by recent findings in neurophysiology. {\sc Spiral}~outperforms the perceptron and {\sc Arow}~by a large margin, when noise is injected at test time, and has lower Rademacher complexity than both of these algorithms. 

\subsubsection*{Acknowledgments}

\bibliographystyle{plainnat}
\bibliography{bibtest}

\end{document}